%% file: ms.tex
\documentclass[twoside]{article}

\usepackage[accepted]{aistats2021}

\input{preamble/packages.tex}
\input{preamble/definitions.tex}

\input{preamble/math.tex}

\begin{document}

\twocolumn[

\aistatstitle{Have We Learned to Explain?: How Interpretability Methods Can Learn to Encode Predictions in their Interpretations.}

\aistatsauthor{ Neil Jethani \And Mukund Sudarshan\And Yindalon Aphinyanaphongs \And Rajesh Ranganath}

\aistatsaddress{ 
NYU Grossman SOM, NYU \\ \href{nj594@nyu.edu}{\texttt{nj594@nyu.edu}}
\And NYU
\And NYU Langone
\And NYU
} ]

\begin{abstract} 
\label{sec:abstract}
\input{sections/abstract}
\end{abstract}

\section{INTRODUCTION}
\label{sec:intro}
\input{sections/introduction}

\section{AMORTIZED EXPLANATIONS}\label{sec:iwfs}
\input{sections/iwfs}

\subsection{Amortized Explanation Methods (AEMs)}\label{sec:aem}
\input{sections/AEM}

\section{PROBLEMS WITH JAMs}\label{sec:problem}
\input{sections/problem}

\section{EVAL-X: THE EVALUATOR}\label{sec:evaluation}
\input{sections/evaluation}

\section{REAL-X, LET US EXPLAIN!}\label{sec:method}
\input{sections/relax}

\section{EXPERIMENTS}\label{sec:experiments}
\input{sections/experiments}

\section{DISCUSSION}\label{sec:discussion}
\input{sections/discussion}

\bibliography{citations}

\newpage
\appendix
\input{supplement}\label{sec:supplement}

\end{document}

%% file: preamble/packages.tex
\usepackage{amssymb, amsmath, amsthm}
\usepackage{xcolor}

\usepackage{microtype}
\usepackage[english]{babel}
\usepackage[parfill]{parskip}
\usepackage{enumitem}
\usepackage{wrapfig}
\usepackage{multicol}
\usepackage{float}
\usepackage{adjustbox}
\usepackage[utf8]{inputenc} % allow utf-8 input

\usepackage[T1]{fontenc}
\usepackage{amsfonts}
\usepackage{bm}
\usepackage{bbm}
\usepackage{times}

\usepackage{fancyhdr}

\usepackage{graphicx}
\usepackage{wrapfig}
\usepackage{tikz}
\usepackage{forest}
\usetikzlibrary{arrows.meta}

\usepackage{booktabs}
\usepackage{caption}

\usepackage[algoruled, algo2e]{algorithm2e}
\usepackage{algorithm}
\usepackage{algorithmic}
\usepackage{listings}
\usepackage{fancyvrb}
\fvset{fontsize=\normalsize}

\usepackage[round]{natbib}

\usepackage[acronym,smallcaps,nowarn]{glossaries}
\glsdisablehyper
\usepackage{hyperref}
\hypersetup{colorlinks=true,linktoc=all}
\usepackage[all]{hypcap}
\hypersetup{citecolor=niceyellow}
\hypersetup{linkcolor=deepblue}
\usepackage{url}
\usepackage[nameinlink]{cleveref}

\usepackage{multirow}
\usepackage{nicefrac}

%% file: preamble/definitions.tex
\newacronym{lime}{lime}{local interpretable model-agnostic explanations}
\newacronym{shap}{shap}{shapley additive explanations}
\newacronym{roc}{roc}{receiver operating characteristic}
\newacronym{auroc}{auroc}{area under the receiver operating characteristic curve}
\newacronym{auprc}{auprc}{area under the precision recall curve}
\newacronym{acc}{acc}{accuracy}
\newacronym{iwfs}{iwfs}{instance-wise feature selection}
\newacronym{aeo}{aeo}{amortized explanation objective}
\newacronym{relax}{REAL-x}{}
\newacronym{realx}{REAL-x}{}
\newacronym{notrelax}{BASE-x}{}
\newacronym{notrealx}{BASE-x}{}
\newacronym{l2x}{L2X}{learning to explain}
\newacronym{invase}{invase}{instance-wise variable selection using neural networks}
\newacronym{aem}{AEM}{amortized explanation method}
\newacronym{cfsr}{cfsr}{control flow feature selection rate}
\newacronym{tpr}{tpr}{true positive rate}
\newacronym{fdr}{fdr}{false discovery rate}
\newacronym{rebar}{rebar}{}
\newacronym{jam}{JAM}{joint amortized explanation method}
\newacronym{evalx}{EVAL-X}{}
\newacronym{evaluator}{EVAL-X}{}

\newtheorem{lemma}{Lemma}

\crefname{thmdef}{definition}{definitions}
\crefname{lemma}{lemma}{lemmas}
\crefname{prop}{proposition}{propositions}
\crefname{algorithm}{algorithm}{algorithms}

\newif\ifprintcomments

\definecolor{deepblue}{HTML}{20639B}
\definecolor{seagreen}{HTML}{3CAEA3}
\definecolor{niceyellow}{HTML}{D19654}

\newlength\myindent
\forestset{
    .style={
        for tree={
            base=bottom,
            child anchor=north,
            align=center,
            s sep+=1cm,
    straight edge/.style={
        edge path={\noexpand\path[\forestoption{edge},thick,-{Latex}] 
        (!u.parent anchor) -- (.child anchor);}
    },
    if n children={0}
        {tier=word, draw, thick, rectangle}
        {draw, diamond, thick, aspect=2},
    if n=1{%
        edge path={\noexpand\path[\forestoption{edge},thick,-{Latex}] 
        (!u.parent anchor) -| (.child anchor) node[pos=.2, above] {Y};}
        }{
        edge path={\noexpand\path[\forestoption{edge},thick,-{Latex}] 
        (!u.parent anchor) -| (.child anchor) node[pos=.2, above] {N};}
        }
        }
    }
}

\makeatletter
\newcommand\footnoteref[1]{\protected@xdef\@thefnmark{\ref{#1}}\@footnotemark}
\makeatother

%% file: preamble/math.tex
\def\eqref#1{equation~\ref{#1}}
\def\1{\bm{1}}

\def\rvr{{\mathbf{r}}}
\def\rvs{{\mathbf{s}}}

\def\rvv{{\mathbf{v}}}

\def\rvx{{\mathbf{x}}}

\def\rvy{{\mathbf{y}}}
\def\rvz{{\mathbf{z}}}

\def\vc{{\bm{c}}}

\def\vr{{\bm{r}}}
\def\vs{{\bm{s}}}

\def\vv{{\bm{v}}}

\def\vx{{\bm{x}}}
\def\vy{{\bm{y}}}
\def\vz{{\bm{z}}}

\DeclareMathAlphabet{\mathsfit}{\encodingdefault}{\sfdefault}{m}{sl}
\SetMathAlphabet{\mathsfit}{bold}{\encodingdefault}{\sfdefault}{bx}{n}

\newcommand{\E}{\mathbb{E}}

\newcommand{\R}{\mathbb{R}}

\DeclareMathOperator*{\argmin}{arg\,min}

\newcommand{\vxi}{\vx^{(i)}}
\newcommand{\vyi}{\vy^{(i)}}

\newcommand{\g}{\, | \,}
\newcommand{\D}{\mathcal{D}}

\DeclareRobustCommand{\ith}[1]{\ensuremath{{#1}^{(i)}}}
\newcommand{\qselector}{q_{\text{sel}}}
\newcommand{\qpredictor}{q_{\text{pred}}}

\newcommand{\qcontrol}{q_{\text{control}}}
\newcommand{\qevaluator}{q_{\text{eval-x}}}

%% file: sections/abstract.tex
While the need for interpretable machine learning has been established, many common approaches are slow, lack fidelity, or hard to evaluate. 
\Acrlongpl{aem} reduce the cost of providing interpretations by learning a global selector model that returns feature importances for a single instance of data.
The selector model is trained to optimize the 
fidelity of the interpretations, as evaluated by a predictor model for the target.
Popular methods learn the selector and predictor model in concert, which we show allows predictions to be encoded within interpretations.
We introduce \acrshort{evalx} as a method to quantitatively evaluate interpretations and \acrshort{realx} as an amortized explanation method, which learn a predictor model that approximates the true data generating distribution given any subset of the input.
We show \acrshort{evalx} can detect when predictions are encoded in interpretations
and show the advantages of \acrshort{realx} through quantitative and radiologist evaluation.

%% file: sections/introduction.tex
The spread of machine learning models within many crucial aspects of society, has made interpretable machine learning increasingly consequential
for trusting model decisions \citep{lipton2017mythos},
identifying model failure modes \citep{zech2018variable}, 
and expanding knowledge \citep{silver2017mastering}.
Interpretability in machine learning is a well-studied problem, and many methods have been introduced to offer an understanding of which features \textit{locally}, in a given instance of data, are important for generating the target.
This goal can be stated as \textit{\acrfull{iwfs}}.
For example, \acrshort{iwfs} produces saliency maps to explains images, where pixels are segmented based on their importance.

Providing interpretable explanations is a difficult problem, and many popular approaches are either slow or lack fidelity and are hard to evaluate.
Locally-linear methods \citep{Lundberg2017, Ribeiro2016} and perturbation methods \citep{Zeiler2014} are slow --- relying on evaluating numerous feature subsets or solving an optimization problem
for each instance of data.
While gradient-based methods \citep{Simonyan2013, Springenberg2014} provide faster explanations, recent studies \citep{adebayo2018sanity, NIPS2019_9167} have shown that their explanations are inaccurate/lack fidelity. 

Recently, multiple works \citep{dabkowski2017real, Chen2018, yoon2018invase, schwab2019cxplain}, which we refer to as \acrfullpl{aem}, amortize the cost of providing model-agnostic explanations by learning a single global selector model that efficiently identifies the subset of locally important features in an instance of data with a single forward pass.
\Acrshortpl{aem} learn the global selector model by optimizing an objective that measures the fidelity of the explanations.

\Acrshortpl{aem} assess feature subset selections, provided as masked inputs, using a predictor model for the target.
Either the original predictor model trained on the full feature set is used or a new predictor model is trained.
If the original predictor model is used and simple masking, such as with a default value, is employed, \citep{dabkowski2017real, schwab2019cxplain} the masked inputs will come from a different distribution than that on which the model was trained, violating a key assumption in machine learning \citep{NIPS2019_9167}.
Instead, \Acrshort{l2x} and \acrshort{invase} fit a new predictor model jointly with the selector model.
We refer to such methods as \acrfullpl{jam}.
\Acrshortpl{jam} have been used for a range of applications --- providing image saliency maps, identifying important sentences in text, and identifying features involved in predicting mortality at the patient-level \citep{Chen2018, yoon2018invase}.

While \acrshortpl{jam} seek to provide users with fast, high fidelity explanations, we show that they encode predictions within interpretations and omit features involved in control flow.
\begin{figure}[t]%{r}{0.3\textwidth}
    % \vspace{-0.2cm}
    \begin{center}
    \includegraphics[width=0.3\textwidth]{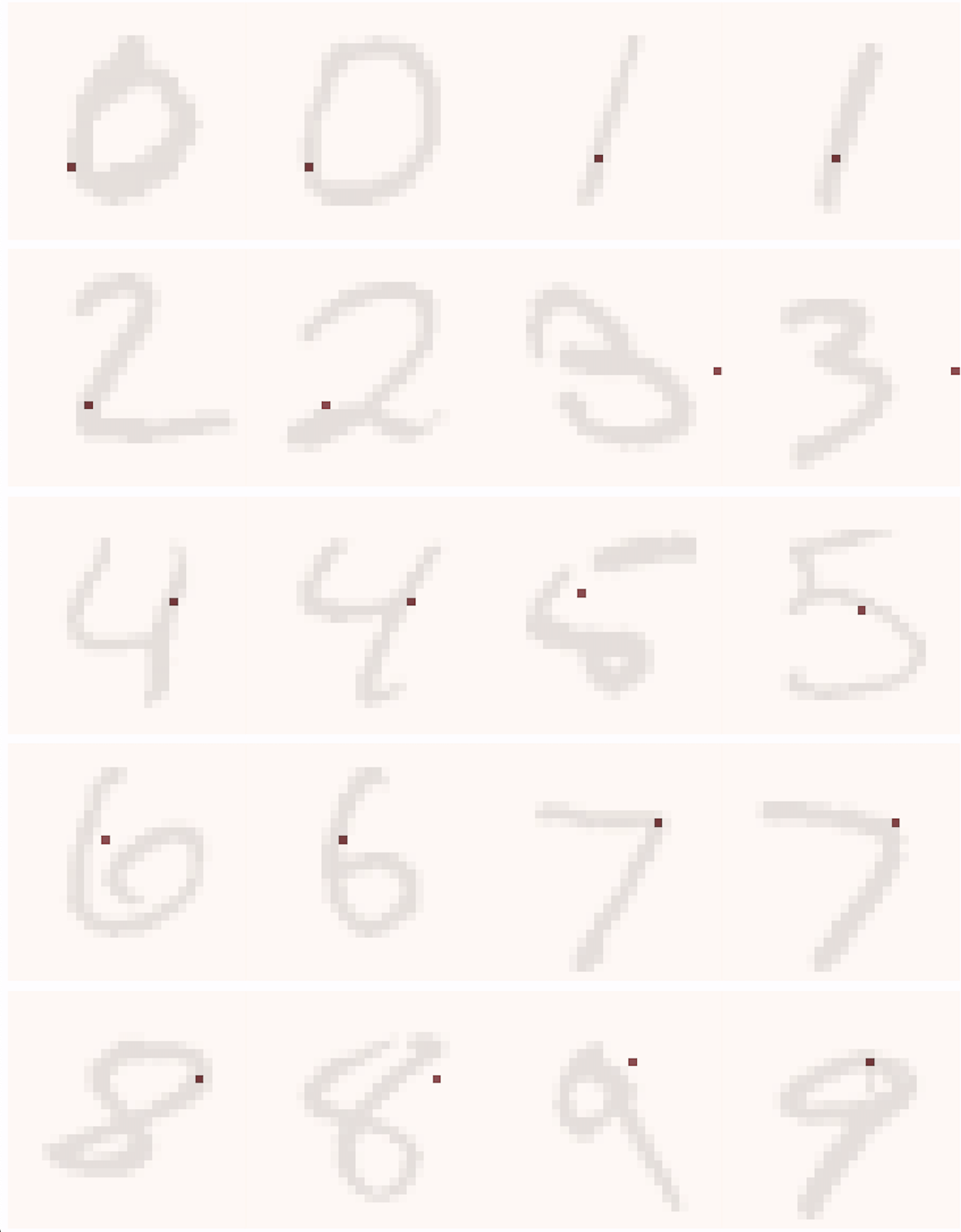}
  \end{center}
  \vspace{-0.2cm}
  \caption{\small {\textbf{\Acrshort{l2x} classifies digits with $96\%$ accuracy from a single feature selection.}
} \normalsize    }
    \label{fig:mnist_1}
    \vspace{-.5cm}
\end{figure}
\Cref{fig:mnist_1} illustrates this point;
it plots explanations from \acrshort{l2x} \cite{Chen2018} on MNIST \citep{LeCun1998} trained with a selector model that outputs a \textit{single} important pixel and achieves 96.0\% accuracy.
Here, the selector model makes the classification decision and transmits it to the predictor model via the binary code of the selections, encoding the prediction.
The issues with \acrshortpl{jam} stem from the predictor model co-adapting to work with the selector model to fit the data, allowing the predictor model to map from selection masks to predictions.
Had the selections in \cref{fig:mnist_1} been evaluated under the true data generating distribution of the target given single feature subsets of the input, the digits could not have been accurately predicted.
To identify such issues in practice, interpretations should be evaluated.

We first develop \acrshort{evalx} 
\footnote{\label{foot:code}Code is available at \href{https://github.com/rajesh-lab/realx}{https://github.com/rajesh-lab/realx}}
as a method to evaluate interpretations, which learns to approximate the true data generating distribution of the target given subsets of the input.
Then, we introduce \acrshort{realx}
\footnoteref{foot:code}
as a novel \Acrshort{aem}, which learns to select minimal feature subsets that maximize the likelihood of the data under an estimate of the true data generating distribution of the target given subsets of the input. 
\Acrshort{realx} provides fast, high fidelity/accuracy explanations without encoding predictions or relying on model predictions generated by out-of-distribution inputs.
We compare \acrshort{relax} to existing \acrshortpl{jam} on established synthetic, MNIST, and Chest X-Ray datasets.
We show that \acrshort{realx} helps address the issues with \acrshortpl{jam} through quantitative \acrshort{evalx} and \textit{expert clinical evaluation.} 
 
\subsection{Related Work}\label{sec:prior_work}

Interpretability methods can be divided into four different approaches -- gradient-based, perturbation-based, locally linear, and \acrlongpl{aem}.

Gradient-based methods, such as \citep{Simonyan2013, Springenberg2014, Shrikumar2017}, calculate the gradient of the target with respect to features in the input.
\citet{Simonyan2013}, for example, does so with imaging data, overlaying a "saliency map" of important pixels. 
Similarly, grad-CAM \citep{Selvaraju_2019} calculates the gradient of the target with respect to intermediate layers in a CNN.
% DeepLIFT \citep{Shrikumar2017}, meanwhile, compares the activations of neurons in a neural network to a reference during backpropagation.
In addition to often requiring strong modeling assumptions (i.e. restricting the model class to CNNs), these methods do not optimize any objective to ensure the fidelity/accuracy of their explanations.  
\citet{NIPS2019_9167} show that the estimates of feature importance derived from many gradient-based methods are often no better than a random assignment of feature importance.
\citet{adebayo2018sanity} 
% attribute the visual appeal of the explanations returned by these methods to the modeling assumptions made by CNNs and 
show that even random model parameters and targets provide seemingly acceptable explanations.

Perturbation-based approaches, such as \citep{Zeiler2014, Zhou2015, Zintgraf2017}, perturb the inputs and observe the effect on the target or neurons within a network in order to gauge feature importance.
This process requires separate forward passes through the network for each perturbation, which is computationally inefficient and can underestimate the importance of features \citep{Shrikumar2017}. 

Locally linear methods, popularly \acrshort{lime} \citep{Ribeiro2016} and \acrshort{shap} \citep{Lundberg2017}, provide an explanation that is a linear function of simplified variables to explain the prediction of a single input.
Locally linear methods assume model linearity in order to explain complex feature interactions and non-linear decision boundaries.
These methods require selecting different feature subsets for each instance in order to assess the their impact on the prediction of the target, a computationally-intensive process. 
In order to assess model predictions given only a subset of features, the missing features are sampled independently, resulting in model inputs that can be out-of-distribution.
This can lead to unexpected results because there is no expectation on what the model will return on out-of-distribution inputs.
While these methods do optimize for the fidelity of their explanations, they are slow, require strong modeling assumptions, such as linearity, and rely on out-of-distribution estimates.  

\Acrlongpl{aem}, \citep{dabkowski2017real, Chen2018, yoon2018invase, schwab2019cxplain}, learn a global model to explain any sample of data.
\Acrshortpl{aem} are the only class of methods that provide both an objective to measure explanation fidelity and fast explanations with a single forward pass. 
\Acrshortpl{aem} accomplish all this while only requiring that the predictor model is differentiable, demanding no strong modeling assumptions.
However, current approaches either rely on out-of-distribution model inputs \citep{dabkowski2017real, schwab2019cxplain} or, as we show, encode predictions \citep{Chen2018, yoon2018invase}.
We address these issues with \Acrshort{realx}.

Meanwhile, evaluating interpretations is less well studied.
Given the interpretations, many approaches mask out the unimportant features in the inputs and assess their ability to predict the target.
Most approaches \citep{samek2017, Chen2018} do so using a prediction model trained on the full feature set.
In this case, the masked inputs come from a different distribution than those on which the model is trained.  
To address this issue, \citet{NIPS2019_9167} introduced ROAR, which instead retrains a prediction model on the masked-inputs.
However, if the predictions are encoded within the interpretations, then ROAR can learn these encodings. 
We introduce \acrshort{evalx} to address these issues. 

%% file: sections/iwfs.tex
We begin by introducing some preliminaries that we refer to throughout the paper.
Let features $\rvx$ be a random vector in $\mathbb{R}^D$, and the response $\rvy \in \{1, \dots, K\}$.
For a given positive integer $j \leq D$, let $\rvx_j$ be the $j$th component of $\rvx$ and $\rvx_\mathcal{S} := \{ \rvx_j \}_{j \in \mathcal{S}}$ be a subset of features, where $\mathcal{S} \subseteq \{ 1, \dots, D \}$.
$F$ is a distribution over $(\rvx, \rvy)$.

For every instance $(\vxi, \vyi) \sim F(\rvx, \rvy)$, \textit{\acrfull{iwfs}} identifies a minimal subset of features $\vxi_{\mathcal{S}^{(i)}}$ that are relevant to the corresponding target $\vyi$, 
Formally, \acrshort{iwfs} seeks $\vxi_{\mathcal{S}^{(i)}}$ such that under the conditional distribution $F(\rvy \mid \cdot)$ \citep{yoon2018invase}
\begin{gather} \label{eqn:iwfs}
    F(\rvy \mid \rvx_{\mathcal{S}^{(i)}} = \vx_{\mathcal{S}^{(i)}}^{(i)}) = F(\rvy \mid \rvx = \vx^{(i)}).
\end{gather}
Here, $F(\rvy \g \rvx)$ can either be the population distribution from which the data is drawn or, to provide model interpretations, a trained model.

%% file: sections/AEM.tex
\Acrshortpl{aem} refer to a general class of interpretability methods that
learn a \textit{global} selector model to identify a subset
of important features \textit{locally} in any given instance of data $(\vxi, \vyi)$.
The selector model is a distribution $\qselector(\rvs \g \rvx ; \beta)$ over a selector variable $\rvs$, which indicates the important features for a given sample of $\rvx$.
For images, the selector model returns the salient pixels.
\Acrshortpl{aem} optimize $\qselector$ with an objective that measures the fidelity of the selections (i.e. the ability of the selections to predict the target).

\paragraph{\Acrfullpl{jam}.}
Recent, popular \acrshortpl{aem}, \acrshort{l2x} \citep{Chen2018} and \acrshort{invase}~\citep{yoon2018invase} learn $\qselector(\rvs \g \rvx ; \beta)$ in concert with a predictor model $\qpredictor(\rvy \g m(\rvx, \rvs) ; \theta)$. 
We refer to such methods as \textit{\acrfullpl{jam}}.
\Acrshortpl{jam} use a regularizer 
$R(\rvs)$ to control the number of selected features and a masking function $m$ to hide the $j$th feature $\vx_j$ with the selector variable $\vs_j$.
For example, the masking function $m$ can replace 
features with a mask token $\texttt{mask}$\footnote{Let $\mathcal{X} = \mathcal{X}_1 \times ... \times \mathcal{X}_D$ 
be a D-dimensional feature space. 
The mask token \texttt{mask} is chosen such that \texttt{mask} is not in in any of the feature spaces $\mathcal{X}_1 \times ... \times \mathcal{X}_D$.} 
using binary indicators $\vs$:
\begin{align}
    m(\ith \vx, \ith \vs)_j = 
    \begin{cases}
        \ith \vx_j & \text{ if } \ith \vs_j = 1\\
        [\texttt{mask}] & \text{ if } \ith \vs_j = 0
    \end{cases}.
    \label{eqn:hard-mask}
\end{align}
To learn the parameters of the selector model, $\beta$, and the predictor model, $\theta$, the \acrshort{jam} objective maximizes
\begin{gather}
    \E_{\vx, \vy \sim F}\E_{\vs \sim \qselector(\rvs \g \vx ; \beta)}\left[\log \qpredictor(\vy \mid m(\vx, \vs) ; \theta) - \lambda R(\vs) \right] \label{eqn:AEM}
\end{gather}
This objective seeks to measure the ability of the selections to predict the target.
\Cref{eqn:AEM} can be optimized
with score function \citep{Glynn1990, Williams1992} or reparameterization gradients \citep{kingma2014autoencoding} and doesn't make any model-specific assumptions like linearity.

\textbf{\Acrshort{invase}} is a \acrshort{jam} that models the selector variable $\rvs$ using independent Bernoulli distributions denoted $\mathcal{B}$
whose probabilities
are given by a function $f$ of the features.
It sets $R(\rvs) = \ell_0(\rvs)$ to enforce sparse feature selections and uses the masking
from \cref{eqn:hard-mask}. 
\Acrshort{invase} also uses $\qcontrol(\vy \mid \vx ; \phi)$ as a control variate within the objective to reduce the variance of the score function gradients during optimization.
The \acrshort{invase} objective for learning $\theta$ and $\beta$ is
\begin{align*}
    \E_{\vx, \vy \sim F}\E_{\vs_j \sim \mathcal{B}(f_{\beta}(\vx)_j)} &\big[\log \qpredictor(\vy \mid m(\vx, \vs);\theta) \\
    &-  \log \qcontrol(\vy \mid \vx ; \phi)  - \lambda \|\rvs\|_0 \big] \label{eqn:INVASE}
\end{align*}
The use of $\qcontrol$ does not alter \Acrshort{invase}'s objective with respect to the 
selector or predictor model, so it fits into the form of \cref{eqn:AEM}.

\textbf{\Acrshort{l2x}} is a \acrshort{jam} that uses $k$ independent samples from a Concrete distribution \citep{Maddison2016, jang2016categorical} to define the selector model in order to make use of reparameterization gradients during optimization. 
In \acrshort{l2x}, $\rvs$ is sampled from $\qselector(\rvs \g \vx ; \beta)$ as
\begin{gather*}
    \vc_j \sim \text{Concrete}(f_{\beta}(\vx)), \quad 
    C = \left[\vc_1, \dots, \vc_k \right] \in \R^{D \times k}, \\
    \vs_i = \max_{1\leq j \leq k} C_{ij} 
\end{gather*}
Selection with the mask function is accomplished with multiplication: $m(\rvx, \rvs) = \rvx \odot \rvs$.
Sparse selections are enforced by selecting a selection limit $k$ that sets the number of samples taken from a Concrete distribution, assigning a hard bound on the number of features selected. 
\Acrshort{l2x} optimizes the following objective for learning $\theta$ and $\beta$:
\begin{gather}
    \E_{\vx, \vy \sim F} \E_{s \sim \qselector(\rvs \g \vx ; \beta)} \left[\log \qpredictor(\vy \mid \vx \odot \vs; \theta)
    \right].
    \label{eqn:L2X} 
\end{gather}
\Cref{eqn:AEM} and \cref{eqn:L2X} represent the same constrained optimization problem, where in \cref{eqn:L2X} the constraint over the number of features selected is applied explicitly via the selection limit $k$.  

%% file: sections/problem.tex
By maximizing an objective for providing high-fidelity selections, \acrshortpl{aem} learn a selector model that makes it fast and simple to explain any new sample of data.
In this section, however, we
reveal a pair of problems with the \acrshort{jam} objective:
\begin{enumerate}[itemsep=-2mm, topsep=0mm, leftmargin=1cm, rightmargin=0cm]
    \item Encoding predictions with the learned selector.
    \item Failure to select features involved in control flow.
\end{enumerate}

\subsection{Encoding Predictions}\label{sec:encoding}

\Acrshortpl{jam} use the selector variable, $\rvs$, to make selections that mask features in the input. 
For simplicity, we focus on the masking function from \cref{eqn:hard-mask} and on independent Bernoulli selector variables $\rvs_j \sim \textrm{Bernoulli}(f_{\beta}(\rvx)_j)$ like in \Acrshort{invase}. 

For noise free classification, the following lemma states that the selector model can encode the target using the selection of at most a single feature in each sample of data. 
The intuition here is that the selector variable $\rvs$ is a binary code that can pass quite a bit of information to predict the target.
A proof is available in \cref{sec:proof-perfect_encoding}.
\begin{lemma}
\label{lem:perfect_encoding}
Let $\rvx \in \R^D$ and target $\rvy \in \{1,...,K\}$.
If $\rvy$ is a deterministic function of $\rvx$ and $K \leq D$, then \acrshortpl{jam} with monotone increasing regularizers $R$ will select at most \textit{one} feature at optimality. 
\end{lemma}
The proof for this lemma works by having the selector $\qselector$ make the classification based on its input $\vx$, encode the class into a binary
code, and transmit this code via the selector variable to $\qpredictor$ while making use of as few bits as possible.
In this setting, if $K \leq D$, selection of only a single feature can encode the target.
Further, if the regularizer $R$ is monotone increasing, then
an encoding with a single feature will be the preferred maximizer of the \acrshort{jam} objective. 

This idea can be generalized to settings where $\rvy$ is not a
deterministic function of $\rvx$.
In this case, levels of uncertainty can be
encoded through the selected features. 
This is formally captured with \cref{lem:noisy_encoding} in \cref{sec:lemmas} and proved in \cref{sec:proof-noisy_encoding}.
Again, the intuition here is that selector variable produces many unique binary combinations. 
Each binary combination of the selector variable acts as an index that the predictor model uses to output a probability vector for the target classes.
When the input dimensionality $D$ is large, a massive number of indices are available.
This allows the selector model to accurately encode uncertainties about the target, without requiring that the predictor model use the input feature values.

\subsection{Omitting Control Flow Features}\label{sec:control_flow}

% \begin{wrapfigure}{r}{0.5\textwidth}
\begin{figure}[H]
    % \vspace{-.5cm}
    \begin{center}
        \input{figures/control_flow}
    \end{center}
    \vspace{0cm}
    \caption{\small {\textbf{Generative process where the $F(\rvy \mid \rvx)$ is a tree.}
    } \normalsize  }
    \label{fig:control_flow}
    \vspace{-.5cm}
% \end{wrapfigure}
\end{figure}
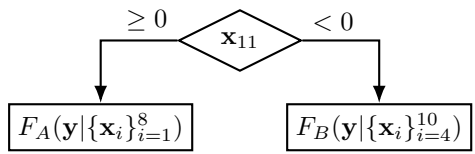

Existing \acrshortpl{aem} can learn to ignore features that only appear in the control flow of a generative process. 
We consider a simple example of such a process in \cref{fig:control_flow}, where $\rvx_i \sim \mathcal{N}(0,1)$. 
Here, $\rvx_{11}$ is involved in a branching decision, such that based on its value either the subset of features $\rvx_{\mathcal{A}} = \{\rvx_i\}_{i=1}^{8}$ or $\rvx_{\mathcal{B}} = \{\rvx_i\}_{i=4}^{10}$ is used to generate $\rvy$. 
We refer to features, like $\rvx_{11}$, that are involved only in the branching decisions/nodes of tree structured generative process as \textit{control flow features}.

Consider using a \acrshort{jam} to explain this process.
In the first case, when $\rvx_{11} \geq 0$, the selector learns to select $\rvx_{\mathcal{A}}$, while the predictor model approximates $F_A$ to generate the target $\rvy$.
Likewise, when $\rvx_{11} < 0$ the selector can select $\rvx_{\mathcal{B}}$ and the predictor can generate $\rvy$ by modeling $F_B$. 
In all cases, the predictor model can use $F_A$ or $F_B$ based on the unique subset of features selected.
\Acrshortpl{jam} do not select $\rvx_{11}$, even though $\rvx_{11}$ is important across all samples. 
\Cref{lem:control_flow} proved in \cref{sec:proof-control_flow} formalizes this phenomenon.

\begin{lemma}
\label{lem:control_flow}
Assume that the true $F(\rvy \mid \rvx)$ is computed as a tree, where the leaves $\ell_i$ are the conditional distributions $F_{i}(\rvy \mid \rvx_{\mathcal{S}_i})$ of $\rvy$ given distinct subsets of features $\mathcal{S}_i$ in $\rvx$.
Given a monotone increasing regularizer $R$, the preferred maximizer of the \acrshort{jam} objective excludes control flow features that are involved in branching decisions. 
\end{lemma}

The proof of this lemma works by having the selector select each distinct subset of features found at the leaves of the tree. 
Given that each distinct subset uniquely maps to an $F_{i}$, the predictor model can learn this mapping and generate the target as well as possible.
Under monotone increasing regularization $R$, the solution that omits control flow features will be preferred over one that selects the full set of relevant features. 
While \acrshort{l2x} does not employ monotone increasing regularization, the \acrshort{l2x} objective still omits control flow features and encodes predictions when the selection limit $k$ is set appropriately to maximize the likelihood of the target while selecting the minimal number of features. 
Both \acrshort{l2x} and \acrshort{invase} benchmark their methods on datasets that contain control flow features. 
We describe these datasets and empirically demonstrate that \acrshortpl{jam} fail to select control flow features in \cref{sec:exp-synth}.

Further, control flow features likely exist in many real-world datasets. 
Consider using electronic health record data to predict mortality in patients presenting with chest pain. 
Troponin lab values, a measure of heart injury, can function as a control flow feature.
Abnormal Troponin values indicate that cardiac imagining should be used to assess disease severity and, therefore, mortality. 
Meanwhile, normal Troponin values indicate that the chest pain may be non-cardiac, and perhaps a chest X-Ray would better inform mortality prediction. 
In this case, using \Acrshortpl{jam} to interpret the prediction will not capture the roll that Troponin plays in determining a patient's mortality.

%% file: figures/control_flow.tex
\forestset{
  default preamble={
    where n children=0{
      tier=word,
    }{
      diamond, aspect=2,
    },
    where level=0{}{
      if n=1{
        edge label={node[pos=.2, above] {$\geq 0$}},
      }{
        edge label={node[pos=.2, above] {$<0$}},
      }
    },
    for tree={
      edge+={thick, -Latex},
      math content,
      s sep'+=1cm,
      draw,
      thick,
      edge path'={ (!u) -| (.parent)},
    }
  }
}

\begin{forest} 
    [\rvx_{11}
        [F_{A}(\rvy|\{\rvx_i\}_{i=1}^8)
        ]   
        [F_{B}(\rvy|\{\rvx_i\}_{i=4}^{10})
        ]   
    ] 
\end{forest}

%% file: sections/evaluation.tex
From the prior sections it is clear that \acrshortpl{jam} can learn to make selections that, instead of selecting the set of relevant features, simply encode their contribution.
In order to trust the explanations provided by \acrshortpl{jam}, selections need to be quantitatively evaluated.

The goal of \acrfull{iwfs} is to find a minimal subset of features $\vxi_{\mathcal{S}^{(i)}}$ that are relevant to the corresponding target $\vyi$, which was stated in \cref{eqn:iwfs} as
$$
F(\rvy \mid \rvx_{\mathcal{S}^{(i)}} = \vx_{\mathcal{S}^{(i)}}^{(i)}) = F(\rvy \mid \rvx = \vx^{(i)}).
$$
The evaluation of \acrshort{iwfs} should reflect the goal---the selections should be evaluated on the \textit{true} conditional distribution $F(\rvy \mid \rvx_{\mathcal{S}^{(i)}} = \vx_{\mathcal{S}^{(i)}}^{(i)})$.
More generally, evaluating any potential selection of a subset of features $\mathcal{R}$ requires access to 
$
F(\rvy \mid \rvx_{\mathcal{R}}).
$
We propose a new method for evaluating \acrshortpl{aem}, called \acrshort{evalx}, which trains an evaluator model $\qevaluator$ to estimate this distribution by maximizing 
\begin{gather} \label{eqn:evaluator}
   \E_{\vx, \vy \sim F}\E_{\vr \sim \mathcal{B}(0.5)}\left[ \log  \qevaluator(\vy \g m(\vx, \vr) ; \eta) \right]. 
\end{gather}
Here $\rvr$ is sampled randomly, independent of $\rvx$, from a Bernoulli distribution, mimicking any potential selection of the input. 
At optimality, \acrshort{evalx} learns the true $F(\rvy \mid \rvx_{\mathcal{R}})$, which we show in \cref{sec:proof-evaluation}.
In practice, reaching optimality may be difficult. 
In \cref{sec:evalx_evaluation} we compare the evaluations returned by \acrshort{evalx} against those returned by distinct models trained on each feature subset and show that \acrshort{evalx} performs similarly, where we consider the set of distinct models as ground truth.
\Cref{alg:evaluator} found in \cref{sec:alg-evaluator} summarizes the \acrshort{evalx} training procedure. 
In practice, predictive performance metrics returned by the \acrshort{evalx} should be used to quantitatively evaluate selections. 
 
Of note, this evaluation differs from the common approach of simply masking the uninformative features and seeing how performance degrades on a model trained on the full feature set.
\citet{Chen2018} suggests evaluating using \textit{post-hoc accuracy} following this approach.
However, as mentioned by \citet{NIPS2019_9167}, samples where a subset of the features are masked are out of the distribution of the original input. 
\citet{NIPS2019_9167} address these out of distribution issues with ROAR, where they suggest training and testing a model on samples from the same distribution of masked inputs, $m(\rvx, \rvs)$.
However, this procedure allows the post-hoc evaluation model to learn the predictions encoded in the selector variable -- precisely what should be avoided. 

Evaluating explanations using \acrshort{evalx} not only aligns with the goal of \gls{iwfs}, but also addresses the out of distribution issues. 

%% file: sections/relax.tex
We now describe our method to ensure that the learned selections also respect the true data distribution given subsets of the input $F(\rvy \g \rvx_{\mathcal{R}})$.

\Acrshortpl{jam} learn to select features and make predictions in concert.
This flexibility allow \acrshortpl{jam} to learn to make predictions from information encoded in the choice of selections.
If the predictor model is learned disjointly, however, this possibility is eliminated. 

Therefore, we propose learning the predictor model disjointly to approximate $F(\rvy \g \rvx_{\mathcal{R}})$ whilst learning to select the minimal subset of features to maximize the probability of the data.
Given the insights of \cref{sec:evaluation}, this procedure is expressed as learning $\qpredictor(\ \cdot\ ; \theta)$ to maximize
\begin{gather*}
   \E_{\vx, \vy \sim F}\E_{\vr \sim \mathcal{B}(0.5)}\left[ \log  \qpredictor(\vy \g m(\vx, \vr) ; \theta) \right],
\end{gather*}
while learning $\qselector(\ \cdot\ ; \beta)$ to maximize
\begin{gather*}
    \E_{\vx, \vy \sim F} \E_{\rvs \sim \qselector(\rvs \mid \vx; \beta)} \left[\log \qpredictor(\vy \mid m(\vx, \vs) ; \theta) - \lambda R(\vs) \right].
\end{gather*}
This modification to the training procedure ensures that $\qpredictor$ respects the true data distribution and avoids encoding predictions within the learned selector variable. 

To make the method concrete, $\qselector$ and $R$
need to be chosen. We introduce the following procedure as \textit{\acrshort{realx}}:
\begin{gather}
    \label{eqn:relax}
    \max_{\beta}   \E_{\vx,\vy} \E_{\rvs_i \sim \mathcal{B}(f_{\beta}(\vx)_i)} \big[\log \qpredictor(\vy \mid m(\vx, \vs);\theta) 
    - \lambda \|\vs\|_0 \big], \nonumber \\ 
    \max_{\theta}   \E_{\vx,\vy} \E_{\vr_i \sim \mathcal{B}(0.5)} \big[\log \qpredictor(\vy \mid m(\vx, \vr);\theta) \big].
\end{gather}
\Acrshort{realx} is a new AEM. 
\Acrshort{realx} learns a global selector model to identify the subset of important features locally in any given instance of data. 
The selector model is trained on a global objective that measures selection fidelity.
\Acrshort{realx} uses discrete selections sampled independently from a Bernoulli distribution and penalizes the number of features selected.

\subsection{Implementation}

To optimize over discrete feature selections, \acrshort{realx} employs \acrshort{rebar} gradients \citep{tucker2017rebar}, a score function gradient estimator that uses relaxed continuous selections within a control variate to lower the variance of the gradient estimates. \Cref{alg:relax} summarizes the training procedure (reference \cref{sec:rebar} for  \cref{eqn:s,eqn:z,eqn:ztilde,eqn:rebar}). 

\vspace{-.1cm}
\begin{algorithm}[H]
\caption{\Acrshort{realx} Algorithm}
\label{alg:relax}
\DontPrintSemicolon
\SetAlgoLined
\KwIn{$\D := (\vx, \vy)$, where $\vx \in \mathbb{R}^{N \times D}$, feature matrix; $\vy \in \mathbb{R}^N$, labels}
\KwOut{$\qselector(\cdot; \beta)$, function that returns feature selections given an instance of $\rvx$}
\textbf{Select: }$\lambda$, regularization constant; $\alpha$, learning rate; $M$, mini-batch size, $T$, training-steps

\For{$1,..., T$}{

Randomly sample mini-batch of size $M$, $( \ith \vx,  \ith \vy)_{i=1}^M \sim \D$

\For{$i = 1,..., M$}{
\textbf{Sample Selections:}

\hskip1.em $\ith \vr \sim \text{Bernoulli}(0.5)$

\hskip1.em Sample $\ith \vs$, $\ith \vz$, and $\Tilde{\vz}^{(i)}$ using $\qselector(\cdot; \beta)$ as in

\hskip14.em  \cref{eqn:s,eqn:z,eqn:ztilde} 
}

\textbf{Optimize Models:}

\hskip.7em $\theta = \theta + \alpha \nabla_{\theta}  \left[ \frac{1}{M} \sum_{i=1}^M \log \qpredictor(\ith \vy | m(\ith \vx, \ith \vr; \theta) \right ]$

\hskip.7em$\beta = \beta + \alpha \frac{1}{M} \sum_{i=1}^M \hat{g_{\beta}}$ \quad ($\hat{g_{\beta}}$ as in \cref{eqn:rebar})
}
\end{algorithm}

\vspace{-.5cm}
Note that the user has the option to train $\qpredictor$ first, then optimize $\qselector$.
We show in \cref{sec:stepwise} that this approach performs similarly. 
The algorithm makes clear that the predictor model 
is updated independently of \acrshort{realx}'s learned selections and, 
therefore, cannot make accurate predictions from 
selections that directly encode the target or 
omit control flow features.

%% file: sections/experiments.tex
We have shown that existing \acrshortpl{aem} can encode predictions within selections and fail to select features involved in control flow.
We then introduced \acrshort{evalx} as a procedure for quantitatively evaluating explanations.
Further, we proposed \acrshort{relax} as a simple method to address the issues with existing \acrshortpl{aem} by mirroring our evaluation procedure. 

In order to properly evaluate \acrshort{relax}, we introduce \acrshort{notrelax} as a baseline to ensure that the results we obtain on \acrshort{relax} are not due to changes in the optimization procedure.
\Acrshort{notrelax} is a \acrshort{jam} that mimics the gradient optimization and regularization procedure of \acrshort{relax}.
We evaluate all the \acrshortpl{jam} --- 
\acrshort{l2x}\footnote{\href{https://github.com/Jianbo-Lab/L2X}{https://github.com/Jianbo-Lab/L2X}},
\acrshort{invase}\footnote{\label{note:invase}\href{https://github.com/jsyoon0823/INVASE}{https://github.com/jsyoon0823/INVASE}}, 
and \acrshort{notrelax} ---
and our method, \acrshort{relax},
on a number of established \textbf{Synthetic Datasets}, \textbf{MNIST}, and on real-world \textbf{Chest X-Rays}.

We show that \acrshort{relax} selects control flow features in synthetic data.
On imaging data, we demonstrate that \acrshort{relax} obtains higher predictive performance on \acrshort{evalx}, while other method's seek to encode the classification. 
Further, we \textit{elicit expert radiologist feedback} to rank the explanations of cardiomegaly returned by each method.

\subsection{Trading-Off Interpretability and Accuracy}
While the goal of \acrshort{iwfs} (\cref{eqn:iwfs}) assumes that a small human-interpretable subset of features generate the target, 
in practice there is a trade-off between interpretability and predictive accuracy. 
This trade-off has been discussed by many prior works \citep{Selvaraju_2019, lakkaraju2017interpretable, ishibuchi2007analysis}.

A \textit{simple} explanation of the data, one with fewer features selected, allows for greater human interpretability.
However, on real-word data this is likely to come at the cost of predictive accuracy.
The \Acrshortpl{aem} considered optimize multiple objectives — an objective that measure the fidelity of the learned selections and an objective that measures the interpretability of the selections by limiting the number of features selected.
By tuning the hyper-parameter balancing these objectives, different solutions along the multi-objective Pareto front can be reached to trade-off interpretability and predictive accuracy.   
For the real-world datasets, we, therefore, choose the hyper-parameter associated with the most interpretable solution such that the following condition is met:
\textit{\Acrfull{acc} is within $5$\% of a model trained on the full feature set.}
We report the performance of the model trained on the full feature set, and refer to this model as FULL.
For our synthetic datasets, we do not need to make this trade-off because we know that only a small number of interpretable features generate the target and instead chose the hyper-parameter that maximizes the accuracy.

\subsection{Evaluation}
We also evaluate the selections obtained by each method using the predictive performance measured by \acrshort{evalx}, which we denote as \textbf{e\acrshort{auroc}} and \textbf{e\acrshort{acc}}.
Together, we show that good predictive performance, as measured by \acrshort{auroc} and \acrshort{acc}, attained by \acrshort{l2x} and \acrshort{invase} does not imply good performance upon evaluation with \acrshort{evalx}. 
A phenomenon that we have described explicitly in \cref{sec:problem}.
Whereas, \acrshort{relax} is more robust to these issues. 

It is possible to use \acrshort{evalx} to select \Acrshort{aem} models.
However, for a \acrshort{jam} that at optimum 
encodes predictions or omits control flow features, the effectiveness of such a procedure would hinge on poor optimization of the \acrshort{jam}'s objective. 

\subsection{Synthetic Datasets}
\label{sec:exp-synth}
Both \acrshort{l2x} and \acrshort{invase} evaluate their methods on a number of synthetic datasets\footnotemark[2], where the data generation procedure is described as follows: 
% \vspace{-.2cm}
\begin{align*}
    {\{\rvx_i\}}_{i=1}^{11} \sim \mathcal{N}(0,1) & \ \ \ \ \ \ \ 
    \rvy \sim \text{Bernoulli}\left(\frac{1}{1+f(\rvx)}\right),
\end{align*}
\vspace{-.2cm}
The functions for $f(\rvx)$ vary as follows:
% \vspace{-.2cm}
% \begin{multicols}{2}
\begin{itemize}[itemsep=-1mm, topsep=0mm]% leftmargin=*, rightmargin=0.5cm]
    \item $f_{\textbf{A}}(\rvx) =$ exp($\rvx_1 \rvx_2$)
    \item $f_{\textbf{B}}(\rvx) =$ exp($\sum_{i=3}^6 \rvx_i^2 -4$)
    \item $f_{\textbf{C}}(\rvx) =$ exp($-10\sin(0.2\rvx_7) + |\rvx_8| + \rvx_9 + e^{-\rvx_{10}}-2.4$),
\end{itemize}
resulting in the following datasets
\begin{itemize}[itemsep=-1mm, topsep=0mm]
    \item \textbf{S1}: If $\rvx_{11}<0$: $f_{\textbf{A}}(\rvx)$; else $f_{\textbf{B}}(\rvx)$
    \item \textbf{S2}: If $\rvx_{11}<0$: $f_{\textbf{A}}(\rvx)$; else $f_{\textbf{C}}(\rvx)$
    \item \textbf{S3}: If $\rvx_{11}<0$: $f_{\textbf{B}}(\rvx)$; else $f_{\textbf{C}}(\rvx)$
\end{itemize}
% \end{multicols}
% \vspace{-.4cm}
This data generating process contains a control flow feature.
Let $\rvx_{\mathcal{A}} = \{\rvx_i\}_{i=1}^2$, $\rvx_{\mathcal{B}} = \{\rvx_i\}_{i=3}^6$, $\rvx_{\mathcal{C}} = \{\rvx_i\}_{i=7}^{10}$, and
$F_J(\rvy \g \rvx) := \text{Bernoulli}\left(\frac{1}{1+f_J(\rvx)}\right)$ for some function $f_J$. $F(\rvy \g \rvx)$ in \textbf{S1-3} is computed as a tree, where $\rvx_{11}$ splits the data into the leaf conditional distributions $F_A(\rvy \g \rvx_{\mathcal{A}})$, $F_B(\rvy \g \rvx_{\mathcal{B}})$, or $F_C(\rvy \g \rvx_{\mathcal{C}})$. 
The feature sets $\rvx_{\mathcal{A}}$, $\rvx_{\mathcal{B}}$, and $\rvx_{\mathcal{C}}$ are distinct and do not contain $\rvx_{11}$. Thus, $\rvx_{11}$ is a control flow feature. 

\paragraph{Model training.}
The training and test sets both contained $10,000$ samples of data.
For all methods, we used neural networks with $3$ hidden layers for the selector model and $2$ hidden layers for the predictor model.
The hidden layers were linear with dimension 200. 
All methods were trained for $1,000$ epochs using Adam for optimization with a learning rate of $10^{-4}$. 
We tuned the hyper-parameters controlling the number of features to select across $k = \{1,2,3,4,5,6,7\}$ for \acrshort{l2x} and $\lambda = \{0.05, 0.075, .1, 0.125, 0.15, .2, .25\}$ for \acrshort{invase}, \acrshort{relax}, and \acrshort{notrelax}.
We select the configuration that yields the largest \acrshort{acc}. 

\input{tables/synth_results}
\paragraph{Results.} We summarize our results for each dataset, paying special attention to the \acrfull{cfsr}, defined by the proportion of control flow features selected. 
We also include the \acrshort{acc}, \acrshort{auroc}, the \acrshort{tpr}, defined by the proportion of important features selected, the \acrshort{fdr}, defined by the proportion of selected features that are not important, and the corresponding \acrshort{evalx} evaluation metrics (e\acrshort{auroc}, e\acrshort{acc}).
We summarize these results in Table \ref{tab:synth_results}, which show that only \acrshort{realx} consistently selects the control flow feature and attains a higher \acrshort{tpr}, e\acrshort{auroc}, and e\acrshort{acc}. 
From \cref{lem:control_flow} in \cref{sec:control_flow} we expect that \acrshortpl{jam} should never select the control flow feature. 
However, the fact that \acrshortpl{jam} occasionally do so is due to either incomplete optimization (\acrshort{invase}) or no preference in selecting the control feature (\acrshort{l2x} when $k$ is large enough).
Yet, requiring \acrshort{relax} to predict well from random selections results in a slightly greater \acrshort{fdr}. 
By not selecting control flow features, \acrshort{l2x} and \acrshort{invase} achieve lesser \acrshortpl{tpr}.  
Though \acrshort{relax} often does not obtain the highest \acrshort{auroc} and \acrshort{acc}, it obtains greater e\acrshort{auroc} and e\acrshort{acc} upon evaluation with \acrshort{evalx} due to the selection of the control flow feature.
These results help highlight \acrshort{relax}'s ability to address issues that prior methods have with selecting control flow features.

\begin{figure*}[t!]
    \centering
    \includegraphics[width=1.\textwidth]{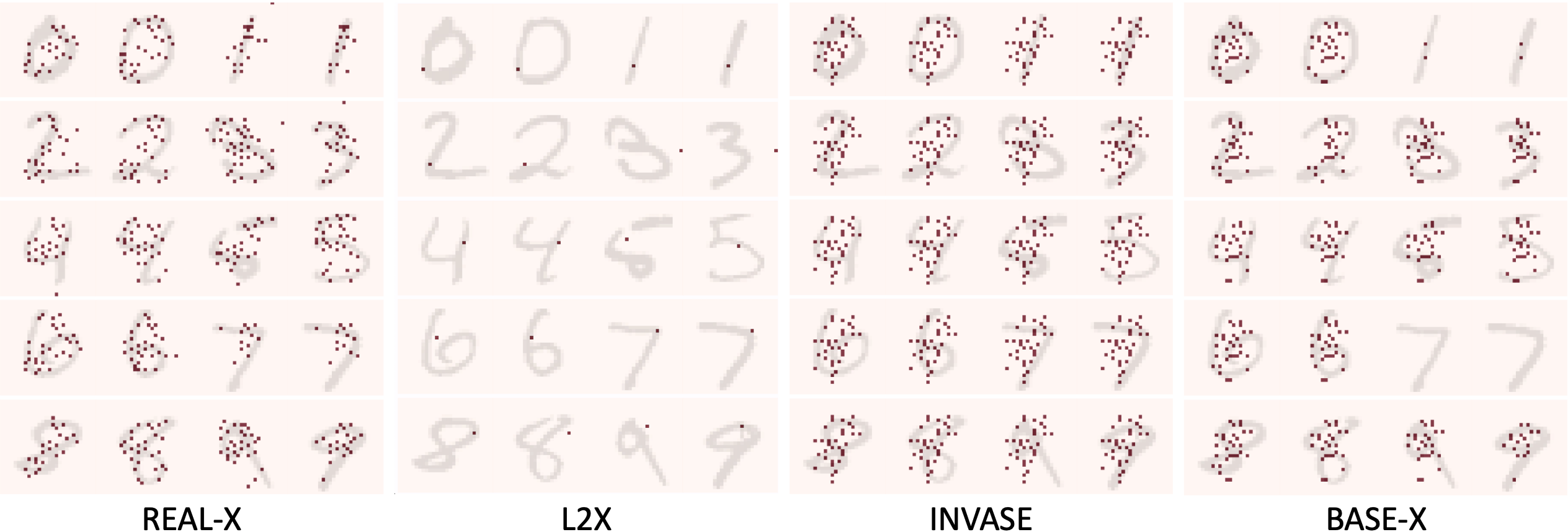}
    \caption{
\small {    \textbf{\acrshort{relax} makes reasonable selections, with other methods encode predictions.}
    Each column is labeled with the method used to learn selections. For each digit two random samples are provided and selections are presented in dark red.
} \normalsize    }
    \label{fig:mnist_results}
\end{figure*}

\begin{figure*}[t!]
    \vspace{0.5cm}
    \centering
    \includegraphics[width=1.\textwidth]{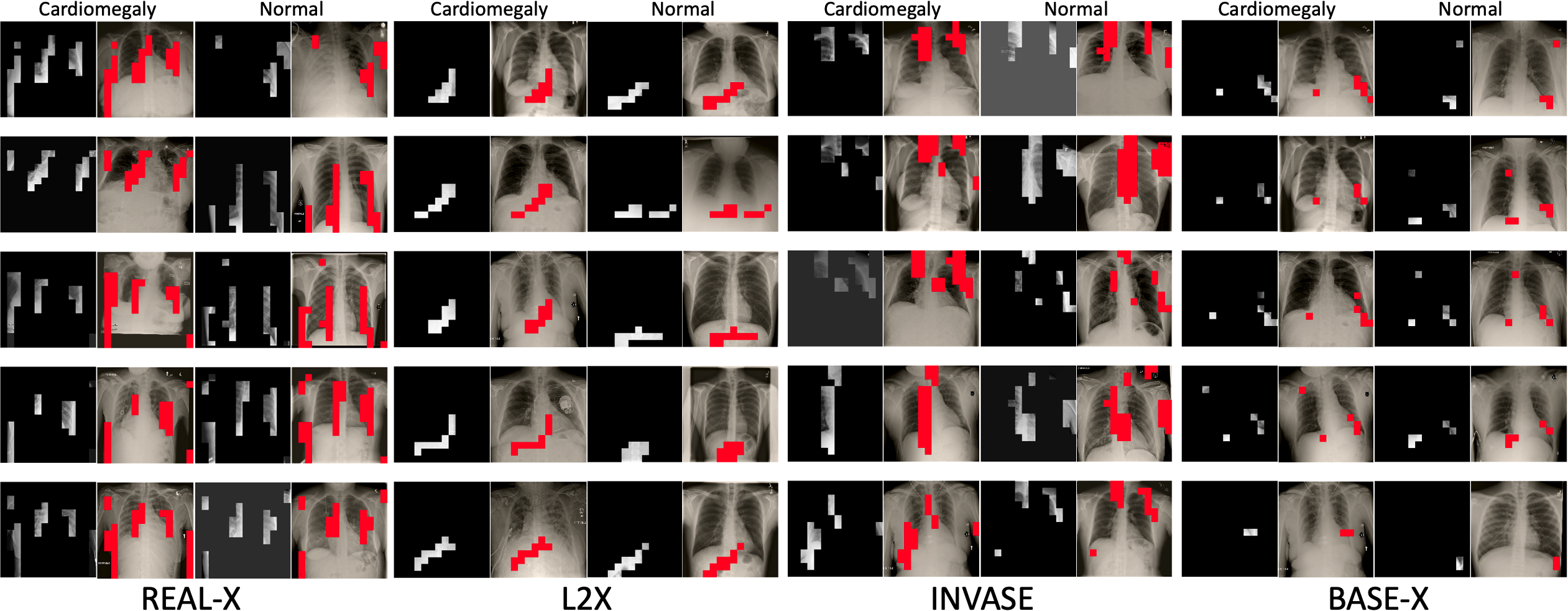}
    \caption{
\small {    \textbf{\acrshort{relax} makes reasonable selections around the margins of the heart without encoding the prediction.}
    5 random samples of Cardiomegaly and Normal Chest X-Rays are presented for each method. The selected inputs are places beside images with the selections overlaid in red. 
} \normalsize    }
    \label{fig:cxr_results}
    % \vspace{-0.2cm}
\end{figure*}
\subsection{MNIST}
The MNIST dataset \citep{LeCun1998} is comprised of $70,000$, $28\times28$ images of handwritten digits in $\{0,...,9\}$.
\paragraph{Model training.}
The images were trained using $60,000$ samples and evaluated on $10,000$ samples.
For all methods, we used neural networks with $3$ hidden layers for the selector model and $2$ hidden layers for the predictor model.
The hidden layers were linear with dimension 200.  
All methods were trained for $500$ epochs using Adam for optimization with a learning rate of $10^{-4}$.
We tuned the hyper-parameters controlling the number of features to select across $k = \{1,5,15,50,100,200\}$ for \acrshort{l2x} and $\lambda = \{0.1, 1.0, 5.0, 10.0, 25.0, 50.0\}$ for \acrshort{invase}, \acrshort{relax}, and \acrshort{notrelax}.
We then selected the configuration that allowed for the smallest number of features to be selected while retaining an \acrshort{acc} within $5$\% of that obtained by a model trained on the full feature set.

\paragraph{Results.} 
\input{tables/mnist_results}
\Cref{tab:mnist_results} shows that while \acrshort{l2x}, \acrshort{invase}, and \acrshort{notrelax} all make selections that allow for high predictive performance (\acrshort{acc} $\geq 92.8\%$), when evaluated with \acrshort{evalx} the predictive performance is significantly worse. 
Looking at the selections made by each method on \cref{fig:mnist_results} helps clarify this discrepancy. 
As we saw earlier, \acrshort{l2x} can encode the prediction with a single feature. 
We also see evidence of encoding with \acrshort{notrelax}, where certain digits, $\{1,7\}$, are encoded by the selection of a few features or no features.
\Acrshort{invase}, however, seems to optimize poorly in high-dimensions and instead selects a conserved set of features across digits.
\acrshort{relax}, however, performs far better upon \acrshort{evalx} evaluation.
Also, \acrshort{relax}'s selections appear reasonable, selecting pixels along the entire digit or along regions that help distinguish digits. 

\subsection{Chest X-Ray Images}

The NIH ChestX-ray8 Dataset\footnote{\href{https://nihcc.app.box.com/v/ChestXray-NIHCC}{https://nihcc.app.box.com/v/ChestXray-NIHCC}} \citep{Wang_2017} contains $112,120$ chest X-ray images from $30,805$ patients, each labeled with the presence of $8$ diseases. 
We selected a subset of $5,600$ X-rays labeled either \textit{cardiomegaly} or \textit{normal}, including all $2,776$ X-rays with cardiomegaly.
Cardiomegaly is characterized by an enlarged heart, and can be diagnosed by measuring the maximal horizontal diameter of the heart relative to that of the chest cavity and assessing the contour of the heart. 
Given this, we expect to see selections that establish the margins of the heart and chest cavity. 

\paragraph{Model training.}
We used $5,000$, $300$, and $300$ images for training, validation, and testing respectively.
UNet and  DenseNet$121$ architectures were used for the selector and predictor models respectively. 
All methods were trained for $50$ epochs using a learning rate of $10^{-4}$.
We choose to learn $16\times16$ super-pixel selections.
We tuned the hyper-parameters controlling the number of features to select across $k = \{1,5,15,50,100\}$ for \acrshort{l2x} and $\lambda = \{0.1, 1.0, 2.5, 5.0, 50.0\}$ for \acrshort{invase}, \acrshort{relax}, and \acrshort{notrelax}.
We then selected the configuration that allowed for the smallest number of features to be selected while retaining an \acrshort{acc} within $5$\% of that obtained by a model trained on the full feature set.

\paragraph{Results.}
\input{tables/cxr_results}
\Cref{tab:cxr_results} shows that while each method makes selections that allow for high predictive performance (\acrshort{acc}$\geq 73.0\%$), \acrshort{relax} yields superior performance upon \acrshort{evalx} evaluation. 
Looking at selections of random Chest X-rays in \cref{fig:cxr_results}, we see that \acrshort{l2x}, \acrshort{notrelax} and \acrshort{invase} seem to make counterintuitive selections that omit many of the important pixels, resulting in a sharp decline in e\acrshort{acc}.

\paragraph{Physician Evaluation.}
\input{tables/cxr_human}
We asked two 
expert
radiologists to rank each method based on the explanations provided.
We randomly selected $50$ Chest X-rays from the test set and displayed the selections made by each method for each X-ray in a random order to each radiologist. 
For a given Chest X-ray, the radiologists then evaluated which selections provided sufficient information to diagnose cardiomegaly and ranked the four options provided, allowing for ties.
In \cref{tab:cxr_human} we report the average rank each method achieved. 
We see that \acrshort{relax} consistently provides explanations that are meaningful to board-certified radiologists.

%% file: tables/synth_results.tex
\begin{table}[H]
\begin{center}
\captionsetup{justification=centering}
\caption{
\small {\textbf{
\Acrshort{realx} achieves superior CFSRs, TPRs, and post-hoc e\acrshort{auroc} on instance-wise tasks.}
} \normalsize}
\vspace{-.1cm}
\tabcolsep=0.3cm
\begin{tabular}{@{}rr|ccc@{}}
\toprule
\rotatebox[origin=c]{0}{Metric}                         &
\rotatebox[origin=c]{0}{Method} & 
\rotatebox[origin=c]{0}{\textbf{S1}} &
\rotatebox[origin=c]{0}{\textbf{S2}} &
\rotatebox[origin=c]{0}{\textbf{S3}} \\ \midrule
\multirow{3}{*}{CFSR}               & \texttt{\Acrshort{realx}}       & \textbf{100.0}            & \textbf{100.0}           & \textbf{100.0}           \\
                                         & \texttt{\Acrshort{l2x}}     & 24.3            & 31.2           & 61.5           \\
                                         & \texttt{\Acrshort{invase}}      & 41.1            & 47.2           & 37.6           \\
                                         & \texttt{\Acrshort{notrealx}}      & 35.0            & 24.5           & 27.2           \\
\midrule
\multirow{3}{*}{TPR}               & \texttt{\Acrshort{realx}}       & \textbf{98.4}            & \textbf{96.7}           & \textbf{93.5}           \\
                                         & \texttt{\Acrshort{l2x}}     & 78.5            & 81.1           & 81.0           \\
                                         & \texttt{\Acrshort{invase}}      & 80.6            & 80.4           & 86.3           \\
                                         & \texttt{\Acrshort{notrealx}}      & 84.7            & 75.6           & 83.5           \\
\midrule
\multirow{3}{*}{FDR}               & \texttt{\Acrshort{realx}}       & 10.7            & 6.3           & 2.6          \\
                                         & \texttt{\Acrshort{l2x}}     & 22.0            & 20.2           & 19.0           \\
                                         & \texttt{\Acrshort{invase}}      & \textbf{1.3}            & 3.1           & \textbf{1.1}           \\
                                         & \texttt{\Acrshort{notrealx}}      & \textbf{1.3}            & \textbf{1.6}           & \textbf{1.1}           \\
\midrule
\multirow{3}{*}{AUROC}               & \texttt{\Acrshort{realx}}       & 0.782            & 0.805           & 0.875           \\
                                         & \texttt{\Acrshort{l2x}}     & 0.752            & 0.790           & 0.852           \\
                                         & \texttt{\Acrshort{invase}}      & \textbf{0.803}            & \textbf{0.806}           & \textbf{0.886}           \\
                                         & \texttt{\Acrshort{notrealx}}      & 0.799           & 0.805           & \textbf{0.886}           \\
\midrule
\multirow{3}{*}{eAUROC}               & \texttt{\Acrshort{realx}}       & \textbf{0.774}            & \textbf{0.804}           & \textbf{0.873}           \\
                                         & \texttt{\Acrshort{l2x}}     & 0.742            & 0.771           & 0.848           \\
                                         & \texttt{\Acrshort{invase}}      & 0.740            & 0.783           & 0.868          \\
                                         & \texttt{\Acrshort{notrealx}}      & 0.762            & 0.773           & 0.867           \\
\midrule
\multirow{3}{*}{ACC}               & \texttt{\Acrshort{realx}}       & 70.1\%            & \textbf{71.5\%}            & 79.6\%           \\
                                         & \texttt{\Acrshort{l2x}}     & 67.1\%           & 70.5\%           & 76.7\%           \\
                                         & \texttt{\Acrshort{invase}}      & \textbf{71.3\%}           & 71.4\%           & \textbf{80.6\%}           \\
                                         & \texttt{\Acrshort{notrealx}}      & 71.0\%           & 71.2\%           & \textbf{80.6\%}           \\
\midrule
\multirow{3}{*}{eACC}               & \texttt{\Acrshort{realx}}       & \textbf{68.6\%}            & \textbf{71.2\%}           & \textbf{79.3\%}           \\
                                         & \texttt{\Acrshort{l2x}}     & 68.4\%            & 70.1\%          & 76.9\%           \\
                                         & \texttt{\Acrshort{invase}}      & 66.8\%            & 69.2\%           & 78.8\%           \\
                                         & \texttt{\Acrshort{notrealx}}      & 68.5\%            & 68.8\%           & 78.7\%           \\
\midrule
\multirow{3}{*}{$k \text{ or } \lambda$}  & \texttt{\Acrshort{realx}}       & 0.05            & 0.05           & 0.1           \\
                                         & \texttt{\Acrshort{l2x}}     & 4            & 4           & 5           \\
                                         & \texttt{\Acrshort{invase}}      & 0.2            & 0.15          & 0.2           \\
                                         & \texttt{\Acrshort{notrealx}}      & 0.15            & 0.125           & 0.2           \\
\bottomrule
\end{tabular}
\label{tab:synth_results}
\end{center}
\vspace{-0.5cm}
\end{table} 

%% file: tables/mnist_results.tex
\begin{table}[h]
\begin{center}
\captionsetup{justification=centering}
\caption{
\small {\textbf{Digit pixels selected by \Acrshort{realx} yield superior results upon post-hoc evaluation.} 
} \normalsize}
\tabcolsep=0.11cm
	\begin{tabular}{r|cc|cc|c}
	\toprule
Method & \acrshort{acc} & \acrshort{auroc} & e\acrshort{acc} & e\acrshort{auroc} & $k\backslash\lambda$ \\ \midrule
\texttt{FULL} & 97.8\% & 0.999 & --- & --- & --\\ 
\midrule
\texttt{\Acrshort{realx}} & 93.8\% & 0.997 & \textbf{86.7\%} & \textbf{0.989} &  5.0 \\
\texttt{\Acrshort{l2x}} & 96.0\% & 0.998 &  11.4\% & 0.561 & 1\\
\texttt{\Acrshort{invase}} & 93.1\% & 0.996 & 50.3\% & 0.883 & 10.0\\
\texttt{\Acrshort{notrealx}} & 92.8\% & 0.996 & 56.9\% & 0.899 & 10.0\\
\bottomrule
\end{tabular}
\label{tab:mnist_results}
\end{center}
\vspace{-0.1cm}
\end{table}

%% file: tables/cxr_results.tex
\begin{table}[H]
\begin{center}
\captionsetup{justification=centering}
\caption{
\small {\textbf{\textbf{\Acrshort{realx} yields superior post-hoc evaluation.}} 
} \normalsize}
\vspace{-.1cm}
\tabcolsep=0.11cm
	\begin{tabular}{r|cc|cc|c}
	\toprule
Method & \acrshort{acc} & \acrshort{auroc} & e\acrshort{acc} & e\acrshort{auroc} & $k\backslash\lambda$ \\ \midrule
\texttt{FULL} & 78.0\% & 0.887 & --- & --- &  -- \\ \midrule 
\texttt{\Acrshort{realx}} & 75.0\% & 0.838 & \textbf{70.3\%} & \textbf{0.777} &  2.5 \\
\texttt{\Acrshort{l2x}} & 75.0\% & 0.848 & 54.0\% & 0.581 &  10 \\
\texttt{\Acrshort{invase}} & 74.3\% & 0.819 & 52.3\% & 0.548 &  2.5 \\
\texttt{\Acrshort{notrealx}} & 74.3\% & 0.818 & 51.7\% & 0.595 &  2.5 \\
\bottomrule
\end{tabular}
\label{tab:cxr_results}
\end{center}
\end{table}
\vspace{-0.4cm}

%% file: tables/cxr_human.tex
\begin{table}[H]
\begin{center}
\captionsetup{justification=centering}
\caption{
\small {\textbf{Average rankings by expert radiologists.} 
} \normalsize}
\vspace{-.1cm}
\tabcolsep=0.11cm
	\begin{tabular}{cccc}
	\toprule
\texttt{\Acrshort{realx}} & \texttt{\Acrshort{l2x}} & \texttt{\Acrshort{invase}} & \texttt{\Acrshort{notrealx}} \\ \midrule
\textbf{1.08 (0.04)} & 3.57 (0.10) & 2.85 (0.11) & 2.29 (0.09) \\
\bottomrule
\end{tabular}
\label{tab:cxr_human}
\end{center}
\end{table}
\vspace{-0.4cm}

%% file: sections/discussion.tex
We proposed \acrshort{realx}, an \acrshort{aem} that provides interpretations that give high likelihood to the data
efficiently with a single forward pass. 
Further, we introduced \acrshort{evalx} as a method to evaluate interpretations, detecting when predictions are encoded in explanations without making out-of-distribution queries of a model. 
\Acrshort{evalx} produces an evaluator model to approximate the true data generating distribution given any subset of the input. 
One future direction could be to produce feature attributions, such as Shapley values, recognizing the evaluator model as a function on subsets for any data point.

With \acrshort{realx}, we employ amortization to provide fast interpretations.
Amortization can help make many existing interpretation techniques scalable, though, as exemplified by \acrshortpl{jam}, care must be taken to avoid encoding predictions within interpretations.
For example, learning a locally linear model to explain each instance of data can be amortized by learning an global explanation model that takes an instance as input and outputs the parameters of a linear model that predicts the target for that instance.
As with \acrshortpl{jam}, the parameters outputted by the explanation model can be used to encode the target.

In addition to extending our methodology to allow for feature attributions, one can explore tailoring it for use with specific data modalities. 
For example, saliency maps are generally more human interpretable if the segmentation is smooth instead of disjoint pixel selections, as in \cref{fig:mnist_results}.
We leave these avenues for future work.

\subsection*{Acknowledgements}

We thank Dr. Lea Azour and Dr. William Moore for clinically evaluating each Chest X-ray explanation.
We also thank the reviewers for their thoughtful feedback.

Neil Jethani was partially supported by NIH T32 GM136573.
Mukund Sudarshan was partially supported by a PhRMA Foundation Predoctoral Fellowship.
Yin Aphinyanaphongs was partially supported by NIH 3UL1TR001445-05 and National Science Foundation award \#1928614.
Mukund Sudarshan and Rajesh Ranganath were partly supported by NIH/NHLBI Award R01HL148248, and by NSF Award 1922658 NRT-HDR: FUTURE Foundations, Translation, and Responsibility for Data Science.

%% file: supplement.tex
\onecolumn
\aistatstitle{Supplementary Materials}

\section{Applying REBAR Gradient Estimation to REAL-X}\label{sec:rebar}
Computing the gradient of an expectation of a function with respect to the parameters of a discrete distribution requires calculating score function gradients.
Score function gradients often have high variance. 
To reduce this variance, control variates are used within the objective.
REBAR gradient calculation involves using a highly correlated control variate that approximates the discrete distribution with its continuous relaxation. 

The \Acrshort{realx} procedure involves
$$
\max_{\beta}   \E_{\vx,\vy} \E_{\rvs_i \sim \mathcal{B}(f_{\beta}(\vx)_i)} \big[\log \qpredictor(\vy \mid m(\vx, \vs);\theta) - \lambda \|\vs\|_0 \big]. 
$$
This is accomplished through stochastic gradient ascent by taking
\begin{gather}\label{eqn:selector_grad}
    \nabla_{\beta} \E_{\rvs_i \sim \mathcal{B}(f_{\beta}(\vx)_i)} \big[\log \qpredictor(\vy \mid m(\vx, \vs);\theta) - \lambda \|\vs\|_0 \big],    
\end{gather}
which requires score function gradient estimation. 

Let $\rvs$ be a discrete random variable, $\mathcal{L} = \E_{\rvs \sim q_{\beta}}[h(\rvs)]$, and $\E[\hat{g}_{\beta}] = \nabla_{\beta}\mathcal{L}$, the REBAR gradient estimator \citep{tucker2017rebar} computes $\hat{g}_{\beta}$. 
Then, letting $\rvz$ be a continous relaxation of $\rvs$, REBAR estimates the gradient as
\begin{gather*}
    \hat{g}_{\beta} = [h(\vs) - h(\Tilde{\vz})]\nabla_{\beta} \log q_{\beta}(\vs) - \nabla_{\beta} h(\Tilde{\vz}) + \nabla_{\beta} h(\vz), \\
    \text{where} \quad \vs =  B(\vz),\ \vz \sim q_{\beta}(\rvz),\ \Tilde{\vz} \sim q_{\beta}(\rvz| \vs).
\end{gather*}

To estimate \cref{eqn:selector_grad} using REBAR, \acrshort{realx} sets
$$
h(\vs) = \log \qpredictor(\vy \mid m(\vx, \ \vs\ );\theta).
$$
Here, $\rvs$ is Bernoulli distributed and \acrshort{realx} sets $\rvz$ to be distributed as the binary equivalent of the Concrete distribution \citep{Maddison2016, jang2016categorical}, which we refer to as the \textit{RelaxedBernoulli} distribution. 
$\rvs$, $\rvz$, and $\Tilde{\rvz}$ are sampled as described by \citet{tucker2017rebar} such that 
\begin{gather}
    p_i = f_{\beta}(\vx)_i, \nonumber \\ 
    \vs_i = B(\vz_i) = \mathbbm{1}(\vz_i > 0), \label{eqn:s}\\
    \vz_i \sim q_{\beta}(\rvz \g \vx) = \textit{RelaxedBernoulli}(p_i; \tau = 0.1), \label{eqn:z}\\
    \Tilde{\rvz_i} \sim q_{\beta}(\rvz \g \vx, \vs) = \frac{1}{0.1} \left(\log \frac{p_i}{1-p_i} + \log \frac{\vv'}{1-\vv'}\right), \label{eqn:ztilde}\\
    \text{ where } 
    \vv \sim \text{Unif}(0,1) \text{ and } 
    \vv' = \begin{cases}
        \rvv(1-p_i) & \text{if }\vs_i=0 \\
        \rvv p_i + (1-p_i) & \text{if }\vs_i=1
    \end{cases}. \nonumber
\end{gather}
Then to estimate \cref{eqn:selector_grad} notice that
$$
\nabla_{\beta} \E_{\rvs_i \sim \mathcal{B}(f_{\beta}(\vx)_i)} \big[\lambda \|\vs\|_0 \big] = \lambda \nabla_{\beta}f_{\beta}(\rvx).
$$
\Acrshort{realx}, therefore, estimates \cref{eqn:selector_grad} by calculating $\hat{g}_{\beta}$ as
\begin{gather}
    \hat{g}_{\beta} = \left[\log \qpredictor(\vy \mid m(\vx, \vs)) - \log \qpredictor(\vy \mid m(\vx, \Tilde{\vz}))\right] \nabla_{\beta} \log \qselector(\vs \g \vx;\beta) - \lambda \nabla_{\beta}f_{\beta}(\vx) \nonumber\\
    -  \nabla_{\beta}\qpredictor(\vy \mid m(\vx, \Tilde{\vz})) +
    \nabla_{\beta}\qpredictor(\vy \mid m(\vx, \vz))
    \label{eqn:rebar}
\end{gather}

\section{Algorithms} \label{sec:alg}

\subsection{Evaluation Algorithm} \label{sec:alg-evaluator}

\begin{algorithm}[H]
\caption{Algorithm to Train Evaluator Model $\qevaluator$}
\label{alg:evaluator}
\DontPrintSemicolon
\SetAlgoLined
\KwIn{$\D := (\vx, \vy)$, where $\vx \in \mathbb{R}^{N \times D}$, feature matrix; $\vy \in \mathbb{R}^N$, labels}
\KwOut{$\qevaluator(\rvy \g m(\vx,\ \cdot\ ); \eta)$, function that returns the probability of the target given a subset of features.}
\textbf{Select: }$\alpha$, learning rate; $M$, mini-batch size

\While{Converge}{

Randomly sample mini-batch of size $M$, $( \ith \vx,  \ith \vy)_{i=1}^M \sim \D$

\For{$i = 1,..., M$}{
\textbf{Sample Selections:}

\hskip1.5em $\ith \vr \sim \text{Bernoulli}(0.5)$
}

\textbf{Optimize:}

\hskip1.5em $\eta = \eta + \alpha \nabla_{\eta}  \left[ \frac{1}{M} \sum_{i=1}^M \log \qevaluator(\ith \vy | m(\ith \vx, \ith \vr); \eta) \right ]$
}
\end{algorithm}

\section{Lemmas}\label{sec:lemmas}

\begin{lemma}
\label{lem:noisy_encoding}
Let $\rvx \in \R^D$, target $\rvy \in \{1,...,K\}$, and $\Delta$ be a set 
of $K$ dimensional probability vectors, then for $J = \argmin_j \sum_{i=0}^j  \genfrac(){0pt}{0}{D}{i} \geq |\Delta|$ and $\rvx \sim F$, there exists a $\qselector$ and $\qpredictor$, where $\{\qpredictor(y = k \g m(\rvx, \rvs))\}_{k=1}^K = \delta(\rvx) \in \Delta$ and $E[||\rvs||_0] \leq J$.
\end{lemma}

\section{Proofs}

\subsection{Proof of Lemma 1}
\label{sec:proof-perfect_encoding}

\paragraph{\Cref{lem:perfect_encoding}.} 
\textit{Let $\rvx \in \R^D$ and target $\rvy \in \{1,...,K\}$.
If $\rvy$ is a deterministic function of $\rvx$ and $K \leq D$, then \acrshortpl{jam} with monotone increase regularizers $R$ will select at most one feature at optimality.} 

As mentioning in \cref{sec:problem}, the lemma considers the masking function from \cref{eqn:hard-mask} and on independent Bernoulli selector variables $\rvs_j \sim \textrm{Bernoulli}(f_{\beta}(\rvx)_j)$.

$\rvs \in \R^D$ is binary and, therefore, has the capacity to transmit D bits of information.
Given that $\rvy \in \{1,...,K\}$ is a deterministic function of $\rvx \in \R^D$, the true distribution is $F(\rvy \mid \vx) \in \{0, 1\}$ for each of the $K$ realizations of $\rvy$. 
Therefore, $m(\vx,\vs)$ must pass at least $\log_2{K}$ bits of information to the predictor model $\qpredictor(\rvy \g m(\vx,\vs))$.  

With $m$ of the form $\cref{eqn:hard-mask}$, this information content can come from $\vs$.
$\vs$ has a capacity of $\log_2\left(\sum_{i=1}^n  \genfrac(){0pt}{0}{D}{i}\right)$ bits when restricted to realizations of $\vs \sim \qselector$ with at most $n$ non-zero elements. 
The maximal number of non-zero elements $J$ in any given realization of $\rvs$ required to minimally transmit $\log_2{K}$ bits of information with $\vs$ can be expressed as 
\begin{align*}
    J = \argmin_j \sum_{i=0}^j  \genfrac(){0pt}{0}{D}{i} \geq K.
\end{align*}
Given $K \leq D$, the maximal number of selections required is given by $J = 1$, where $\genfrac(){0pt}{0}{D}{1} = D \geq K$.
Therefore there exists a $\qpredictor$ and $\qselector$ such that $\E[\qpredictor(\vy \g m(\vx, \vs))] = \E[F(\rvy \g \rvx)]$ and $\E[\|\rvs\|_0] \leq 1$.
For monotone increasing regularizer $R$, any solution that selects more than a single feature will have a lower \acrshort{jam} objective. 
Therefore, at optimally, \acrshortpl{jam} will select at most a single feature.

\subsection{Proof of Lemma 3}
\label{sec:proof-noisy_encoding}

\paragraph{\Cref{lem:noisy_encoding}.} 
\textit{Let $\rvx \in \R^D$, target $\rvy \in \{1,...,K\}$, and $\Delta$ be a set 
of $K$ dimensional probability vectors, then for $J = \argmin_j \sum_{i=0}^j  \genfrac(){0pt}{0}{D}{i} \geq |\Delta|$ and $\rvx \sim F$, there exists a $\qselector$ and $\qpredictor$, where $\{\qpredictor(y = k \g m(\rvx, \rvs))\}_{k=1}^K = \delta(\rvx) \in \Delta$ and $E[||\rvs||_0] \leq J$.}

This proof follows from the proof in \cref{sec:proof-perfect_encoding}. 
Given $\rvx \in \R^D$ and target $\rvy \in \{1,...,K\}$, 
there exists a distribution $\qpredictor(\rvy \g m(\rvx, \rvs))$ such that each realization of $\rvs \in \{0,1 \}^D$ has a bijective mapping to a unique probability vector obtained as $\left\{\qpredictor(y = k \g m(\rvx, \rvs))\right\}_{k=1}^K \in \R^K$. 

As stated in the proof of \cref{lem:perfect_encoding} $\rvs$ has a capacity of $\log_2\left(\sum_{i=1}^n  \genfrac(){0pt}{0}{D}{i}\right)$ bits when restricted to realizations of $\rvs \sim \qselector$ with at most $n$ non-zero elements.
Given a set of $K$ dimensional probability vectors $\Delta$,
the maximal number of non-zero selections in $\rvs$ required to produce at least $|\Delta|$ unique realizations of $\rvs$, denoted by $J$, can be expressed as 
$$
J = \argmin_j \sum_{i=0}^j  \genfrac(){0pt}{0}{D}{i} \geq |\Delta|.
$$
Then there exists a $\qpredictor$ and $\qselector$ such that there are at least $|\Delta|$ unique probability vectors $\left\{\qpredictor(y = k \g m(\rvx, \rvs))\right\}_{k=1}^K = \delta(\rvx) \in R^K$ where $\delta(\rvx) \in \Delta$ and the average number of features selected $E[||\rvs||_0] \leq J$.

\subsection{Proof of Lemma 2}
\label{sec:proof-control_flow}

\paragraph{\Cref{lem:control_flow}.} 
\textit{
Assume that the true $F(\rvy \mid \rvx)$ is computed as a tree, where the leaves $\ell_i$ are the conditional distributions $F_{i}(\rvy \mid \rvx_{\mathcal{S}_i})$ of $\rvy$ given distinct subsets of features $\mathcal{S}_i$ in $\rvx$.
Given a monotone increasing regularizer $R(|\mathcal{S}_i|)$, the preferred maximizer of the \acrshort{jam} objective excludes control flow features.
}

The main intuition behind the proof of \cref{lem:control_flow} is as follows.
The \acrshort{jam} objective results in a prediction model that does not require the control flow features to achieve optimal performance.
As a result of the monotone increasing regularizer $R$, which assigns a cost for selecting each additional feature, the JAM objective omits control flow features.
We now prove this idea formally.

The tree is structured such that each leaf $\ell_i$ in the tree has a corresponding
conditional distribution $F_{i}(\rvy \mid \rvx_{\mathcal{S}_i})$ parameterized by a set of features $\mathcal{S}_i$ such that $\forall j\neq i$, $\mathcal{S}_i \neq \mathcal{S}_j$.
Let the features found along the the path from the root of the tree to the leaf, including those found at the leaf, be defined as $\mathcal{T}_i$ for each leaf $\ell_i$.
Those features that are not in the leaf and only appear in the non-leaf nodes of the tree are the control flow features defined as $\mathcal{C}_{i} := \mathcal{T}_{i} \backslash \mathcal{S}_i$.
For any input $\vx$, let $\mathcal{T}(\vx)$ be the features found along the path used in generating the response for that $\vx$ and define the control flow features along the path as $\mathcal{C}(\vx)$ and set of leaf features $\mathcal{S}(\vx)$.

Consider the following cases where $\qselector(\rvs \g \vx)$ selects the $j$th feature with probability
\begin{align*}
    \begin{cases}
        q_{\text{sel1}}(\rvs_j \g \vx) = \mathbbm{1}[j \in \mathcal{T}(\vx)] = \mathbbm{1}[j \in \{\mathcal{C}(\vx) \cup \mathcal{S}(\vx)\}]  & (\text{Case 1})\\
        q_{\text{sel1}}(\rvs_j \g \vx) = \mathbbm{1}[j \in \mathcal{S}(\vx)]& (\text{Case 2})\\
    \end{cases},
\end{align*}
where $q_{\text{sel1}}$ and $q_{\text{sel1}}$ denotes the $\qselector$ for case 1 and case 2 respectively.
$q_{\text{pred1}}(\rvy \mid m(\vx, \vs))$ and $q_{\text{pred2}}(\rvy \mid m(\vx, \vs))$ are defined in the corresponding manner. 

In case 1, the predictor model $q_{\text{pred1}}$ receives all the relevant features from $q_{\text{sel1}}$, such that $\E_{\vx,\vy \sim F}\E_{\vs \sim q_{\text{sel1}}(\rvs \g \vx)} \left[\log q_{\text{pred1}}(\rvy \mid m(\vx, \vs)) \right]$ can predict as well as possible.

In case 2, however, the predictor model $q_{\text{pred2}}(\rvy \mid m(\vx, \vs))$
does not receive the full set of relevant features from $q_{\text{sel2}}$; it only receives the leaf features. 
Since the leaf features are unique across leaves, the selections indicated by $\vs$ provides enough information for the predictor model to consistently learn the correct data generating leaf conditional $F(\rvy \g \vx_{\mathcal{S}(\vx)})$, meaning that it can predict 
as well as possible.

Assuming the models maximize the \acrshort{jam} objective, in both cases $\qpredictor$ together with $\qselector$ correctly model $F(\rvy \mid \rvx)$.
Plugging this information into the \acrshort{jam} objective in \cref{eqn:AEM} yields the following: 
\begin{align*}
    \mathcal{L}_{\text{Case 1}} &= \E_{\vx,\vy \sim F}\E_{\mathcal{T}(\vx) \sim q_{\text{sel1}}(\rvs \g \vx)} \left[\log q_{\text{pred1}}(\vy \mid m(\vx, \mathcal{T}(\vx))) - \lambda R(|\mathcal{T}(\vx)|)\right] \\
    &= \E_{\vx,\vy \sim F}[\log F(\vy \g \vx)] - \lambda \E_{\vx,\vy \sim F}\E_{\mathcal{T}(\vx) \sim q_{\text{sel1}}(\rvs \g \vx)}[ R(|\mathcal{T}(\vx)|)], \\
    \mathcal{L}_{\text{Case 2}} &= \E_{\vx,\vy \sim F}\E_{\mathcal{S}(\vx) \sim q_{\text{sel2}}(\rvs \g \vx)} \left[\log q_{\text{pred2}}(\vy \mid m(\vx, \mathcal{S}(\vx))) - \lambda R(|\mathcal{S}(\vx)|)\right] \\
    &= \E_{\vx,\vy \sim F}[\log F(\vy \g \vx)] - \lambda \E_{\vx,\vy \sim F}\E_{\mathcal{S}(\vx) \sim q_{\text{sel2}}(\rvs \g \vx)}[ R(|\mathcal{S}(\vx)|)]. 
\end{align*}
Given that $R(.)$ is monotone increasing, the following inequality holds:
\begin{align*}
    \mathcal{L}_{\text{Case 2}} \geq \mathcal{L}_{\text{Case 1}}.
\end{align*}
For any $\lambda > 0$ where control flow features are involved in the data generating process, that is $\mathcal{C}_i \neq \emptyset$ for some $i$, this inequality is strict. 
Therefore, the solution that omits control flow features ($\text{Case 2}$) will have a higher objective value, which we describe as the preferred maximizer of the \Acrshort{jam} objective.
Thus, at optimality, control flow features will not be selected under the \Acrshort{jam} objective with a monotone increasing regularizer.

\section{Optimality of the Evaluator Model}
\label{sec:proof-evaluation}
The evaluator model $\qevaluator$ is learned such that \cref{eqn:evaluator} is maximized as follows:
$$
    \max_{\eta} \E_{\vx, \vy \sim F}\E_{\vr_i \sim \text{Bernoulli}(0.5)}\left[ \log  \qevaluator(\vy \g m(\vx, \vr);\eta) \right]. 
$$
We aim to show that this expectation is maximal when $\qevaluator(\vy \g m(\vx, \vr)) = F(\vy \g \vx_{\mathcal{R}})$ for any sample of $\rvr$ identifying the corresponding subset of features $\mathcal{R}$ in the input $\vx_{\mathcal{R}}$.

The expectations can be rewritten as
$$
\max_{\eta} \E_{\vr_i \sim \text{Bernoulli}(0.5)}\E_{\vx, \vy \g \vr \sim F}\left[ \log  \qevaluator(\vy \g m(\vx, \vr);\eta) \right].
$$
Let the power set over feature selections $\mathcal{P}_{\vr} = \{\vr \subset \{0,1\}^D\}$ and equivalently for the corresponding feature subsets $\mathcal{P}_R = \{\mathcal{R} \subset 2^D\}$. 
Given $\vr_i \sim \text{Bernoulli}(0.5)$, the probability
$$
p(\vr) = \frac{1}{|\mathcal{P}_{\vr}|} = \frac{1}{|\mathcal{P}_R|}.
$$
Recognizing that $\rvx, \rvy \perp \rvr$, the expectation over $\rvr$ can be expanded as
$$
\max_{\eta} \sum_{\vr \in \mathcal{P}_{\vr}} \frac{1}{|\mathcal{P}_{\vr}|} \E_{\vx, \vy \sim F }\left[ \log  \qevaluator(\vy \g m(\vx, \vr);\eta) \right].
$$
Here, the expectation is with respect to a given $\vr$ in the power set $\mathcal{P}_{\vr}$.
In this case, neither $\vr$ nor the subset of features masked by $m(\vx, \vr)$ provide any information about the target.
Therefore, the likelihood is calculated with respect to the corresponding fixed subset $\mathcal{R}$ as 
$$
\max_{\eta} \sum_{\mathcal{R} \in \mathcal{P}_R} \frac{1}{|\mathcal{P}_R|} \E_{\vx, \vy \sim F}\left[ \log  \qevaluator(\vy \g \vx_{\mathcal{R}};\eta) \right].
$$
A finite sum is maximized when each individual element in the sum is maximized, therefore it suffices to find
$$
\max_{\eta} \E_{\vx, \vy \sim F}\left[ \log  \qevaluator(\vy \g \vx_{\mathcal{R}};\eta) \right] \quad \forall \mathcal{R}\in \mathcal{P}_R
$$
Let $\qevaluator := \{f_{\mathcal{R}}(\ \cdot\ ;\eta_{\mathcal{R}})\}_{{\mathcal{R}} \in \mathcal{P}_R}$, such that when given $\vr$ as an input for the corresponding $\mathcal{R}$, $f_{\mathcal{R}}(\ \cdot\ ;\eta_{\mathcal{R}})$ is used to generate the target.
The key point here is that the subset $\mathcal{R}$ provided to the model as $\vr$ can uniquely identify which $f_{\mathcal{R}}$ generates the target.
Then, for any given $R$, each expectation is maximized when the corresponding $f_{\mathcal{R}}$ is equal to the true data generating distribution given by  
$$
\max_{\eta} \E_{\vx, \vy \sim F}\left[ \log  \qevaluator(\vy \g \vx_{\mathcal{R}};\eta) \right] = \max_{\eta_{\mathcal{R}}} \E_{\rvx, \rvy}\left[ \log  f_{\mathcal{R}}(\rvy \g \rvx_{\mathcal{R}} ;\eta_{\mathcal{R}}) \right] = \E_{\rvx, \rvy}[\log  F(\rvy \g \rvx_{R})] \quad \forall \mathcal{R}\in \mathcal{P}_R.
$$

\newpage
\section{Additional Experiments}
\label{sec:additional_experiments}

\subsection{EVAL-X vs. Models Explicitly Trained For Each Feature Subset.}
\label{sec:evalx_evaluation}

In this experiment, we evaluate \Acrshort{evalx}. 
\Acrshort{evalx} approximates $F(\rvy \mid \rvx_{\mathcal{R}})$ for any subset of features $\mathcal{R}$, by training on randomly sampled subsets of the input $\rvx$.
While, at optimally the training procedure for \Acrshort{evalx} returns a model of $F(\rvy \mid \rvx_{\mathcal{R}})$, it may be difficult to approximate the distribution $F(\rvy \mid \rvx_{\mathcal{R}})$ for every possible subset of features. 
We therefore trained separate models for each unique subset of features on our synthetic dataset described in \cref{sec:exp-synth}. 
Each of the datasets contain $11$ input features with $2048$ distinct feature subsets. 
For each dataset we trained $2048$ distinct models on each feature subset. 
We then evaluated the selections made by each \acrshort{aem} on this collection of models and on \acrshort{evalx}.
We compared the \acrshort{auroc} returned by \acrshort{evalx} (e\acrshort{auroc}) to those returned by the collection of models (c\acrshort{auroc}) in \cref{tab:evalx_evaluation}.
While \acrshort{evalx} returns underestimates relative to the collection of models, the difference is small and the trend amongst methods is conserved. 

\input{tables/evalx_evaluation}

\subsection{Training the Predictor Model First.}
\label{sec:stepwise}

We compared a $\acrshort{realx}$ approach where $\qpredictor$ is first fully optimized, then $\qselector$ is optimized in a stepwise manor (\acrshort{realx}-STEP) to the approach outlined in \cref{alg:relax}, where both $\qselector$ and $\qpredictor$ are optimized simultaneous with each mini-batch. 
The \acrshort{auroc}s returned by \acrshort{evalx} for the synthetic datasets described in \cref{sec:exp-synth} are presented in \cref{tab:stepwise}.
Both approaches perform similarly. 

\input{tables/stepwise}

%% file: tables/evalx_evaluation.tex
\begin{table}[H]
\begin{center}
\captionsetup{justification=centering}
\caption{
\small {\textbf{\textbf{\Acrshort{realx} yields superior post-hoc evaluation on a collection of each models for each feature subset.}} 
} \normalsize}
\vspace{-.1cm}
\tabcolsep=0.11cm
	\begin{tabular}{r|cc|cc|cc}
	\toprule
    & \multicolumn{2}{c}{S1} & \multicolumn{2}{c}{S2} & \multicolumn{2}{c}{S3} \\ \midrule
Method & e\acrshort{auroc} & c\acrshort{auroc} & e\acrshort{auroc} & c\acrshort{auroc} & e\acrshort{auroc} & c\acrshort{auroc} \\ \midrule
\texttt{\Acrshort{realx}}       & 0.774    &   \textbf{0.798}    & 0.804      &  \textbf{0.807}   & 0.873    &    \textbf{0.876}   \\
\texttt{\Acrshort{l2x}}     & 0.742      &   0.759   & 0.771      &  0.776   & 0.848      &     0.849\\
\texttt{\Acrshort{invase}}      & 0.740    &   0.767     & 0.783     &   0.788   & 0.868     & 0.870    \\
\texttt{\Acrshort{notrealx}}      & 0.762    &  0.773     & 0.773    &   0.777    & 0.867     & 0.870  \\
\bottomrule
\end{tabular}
\label{tab:evalx_evaluation}
\end{center}
\end{table}
\vspace{-0.4cm}

%% file: tables/stepwise.tex
\begin{table}[H]
\begin{center}
\captionsetup{justification=centering}
\caption{
\small {\textbf{\textbf{\Acrshort{realx} and \Acrshort{realx}-STEP perform similarly.}} 
} \normalsize}
\vspace{-.1cm}
\tabcolsep=0.11cm
	\begin{tabular}{r|cc|cc|cc}
	\toprule
    & \multicolumn{2}{c}{S1} & \multicolumn{2}{c}{S2} & \multicolumn{2}{c}{S3} \\ \midrule
Metric & \texttt{\acrshort{realx}} &  \texttt{\acrshort{realx}-STEP} & \texttt{\acrshort{realx}} &  \texttt{\acrshort{realx}-STEP} & \texttt{\acrshort{realx}} &  \texttt{\acrshort{realx}-STEP} \\ \midrule
e\acrshort{auroc}        & 0.774    &   0.778    & 0.804      &  0.801   & 0.873    &    0.872   \\
\bottomrule
\end{tabular}
\label{tab:stepwise}
\end{center}
\end{table}
\vspace{-0.4cm}